\documentclass[runningheads]{llncs}

\usepackage{booktabs}
\usepackage{multirow}
\usepackage[T1]{fontenc}
\usepackage{amsmath} 
\usepackage{amssymb}
\usepackage[
    colorlinks=true,
    citecolor=blue,
    linkcolor=blue,
    urlcolor=blue
]{hyperref}
\usepackage{caption}

\usepackage{graphicx,verbatim}
\usepackage{color}

\begin{document}
\title{3D Masked Autoencoders are Robust Learners of Volumetric and Multimodal Cellular Representations for Microscopy}

\titlerunning{3D Masked Autoencoders for Cellular Representation Learning}



%
\author{Amirhossein Kardoost\inst{1,5}
\and
Lion Gleiter\inst{1} \and
Tingying Peng\inst{1} \and \\
Carsten Marr\inst{1,2,3,4,5}}

%
\authorrunning{A. Kardoost et al.}
%
\institute{Institute of AI for Health \& Helmholtz AI, Computational Health Center, Helmholtz Munich – German Research Center for Environmental Health, Neuherberg, Germany \and
Department of Medicine III, Ludwig-Maximilian-University Hospital, Munich, Germany \and
Department of Physics, Ludwig-Maximilian-University, Munich, Germany \and
German Cancer Consortium (DKTK), partner site Munich, Germany \and
Munich Center for Machine Learning (MCML), Munich, Germany 
}


\authorrunning{A. Kardoost et al.}  

\maketitle              
\begin{abstract}

Self-supervised learning in fluorescence microscopy often relies on 2D projections, despite the inherently three-dimensional nature of cells. We present a systematic comparison of 2D and 3D masked autoencoders (MAE-2D vs. MAE-3D) on volumetric microscopy data. Under matched architectures and training protocols, MAE-3D consistently outperforms 2D max-projection and slice-based variants on downstream single-cell tasks. We further align visual representations with a pretrained protein language model (ESM2) and show that cross-modal supervision yields larger gains for volumetric models. Channel cross-attention and frequency-domain regularization are critical for leveraging 3D spatial context. On protein--protein interaction prediction, our best model achieves a ROC--AUC of 0.86, while on protein localization it reaches an AUC$_{\text{micro}}$ of 0.95 and an F1$_{\text{micro}}$ of 0.74, demonstrating competitive performance on both tasks. Overall, our findings highlight the potential of volumetric modeling and multimodal alignment for representation learning in single-cell microscopy.


\end{abstract}

\begingroup
\renewcommand{\thefootnote}{}
\footnotetext{Corresponding authors:\\
\{amirhossein.kardoost, tingying.peng, carsten.marr\}@helmholtz-munich.de}
\endgroup
\clearpage
\section{Introduction}

Cells constitute the fundamental building blocks of tissues and organs. Their function is tightly linked to subcellular structure and spatial organization. Understanding cellular organization remains a central challenge in biology. Despite extensive studies~\cite{CAMAE,CellDINO,DINO4Cell,Subcell,ChAdaViT}, deciphering subcellular architecture and protein localization remains complex, particularly in high-dimensional imaging data. Fluorescence microscopy~\cite{OpenCell,JUMP,WTC11} enables visualization of intracellular structures by tagging proteins and organelles with fluorescent markers. Large-scale resources such as JUMP~\cite{JUMP}, OpenCell~\cite{OpenCell}, WTC-11~\cite{WTC11}, and the Human Protein Atlas (HPA)~\cite{HPA} provide multi-channel imaging data capturing rich subcellular organization. Notably, OpenCell and WTC-11 consist of volumetric $z$-stacks. However, many representation learning approaches such as Subcell~\cite{Subcell} and DINO4Cell~\cite{DINO4Cell} operate on 2D projections of these volumes, discarding depth-resolved structural information. We investigate the role of volumetric modeling for the learning of cellular representations. On OpenCell~\cite{OpenCell}, we systematically compare 2D and 3D masked autoencoder (MAE)~\cite{SelfMedMAE,CAMAE} models. We demonstrate that preserving full 3D structure yields more informative representations and consistently improves downstream performance compared to 2D max-projection and even slice-based inputs. Beyond purely visual modeling, we explore multimodal integration by incorporating protein sequence information via a pretrained protein language model (PLM) such as ESM2~\cite{ESM2}. By aligning image features with protein embeddings, we infuse the representation space with biologically grounded structural priors. We  demonstrate that sequence-level supervision enhances representation quality, particularly when coupled with volumetric modeling.

Our contributions are threefold:
(1) We demonstrate that 3D MAE models outperform 2D counterparts across two downstream tasks. 
(2) We show that channel cross-attention and frequency-domain (FFT) regularization further enhance volumetric representation learning.
(3) We establish that integrating protein language models into the visual framework improves representation quality and downstream performance, highlighting the benefit of multimodal alignment for cellular imaging. Code is available at \url{https://github.com/marrlab/mae3d-opencell}.

\section{Related Work}

Fluorescence imaging of proteins, combined with DNA or membrane reference markers, enables single-cell analysis of protein localization and function~\cite{OpenCell,HPA,Cytoself}. OpenCell~\cite{OpenCell} contains 1,310 endogenously tagged human proteins and 29,922 experimentally measured protein--protein interactions, acquired as high-resolution 3D $z$-stacks. Its protein diversity and volumetric imaging make it well suited for studying 3D representation learning and integration of sequence-level embeddings that encode structural information.
The WTC-11 dataset~\cite{WTC11} contains 3D fluorescence images of 25 endogenously tagged proteins (cellular structures), captured with DNA and membrane reference channels and supports tasks such as protein localization and cell cycle stage classification. Compared to OpenCell, WTC-11 exhibits substantially lower protein diversity (25 vs.\ 1,310 proteins), focusing instead on detailed structural characterization across single cells.
Cytoself~\cite{Cytoself} trains a vector-quantized variational autoencoder~\cite{VQVAE} on 2D max-projection images with protein identity conditioning, demonstrating that the learned representations cluster according to subcellular localization.
DINO4Cell~\cite{DINO4Cell} applies self-supervised DINO training~\cite{DINO} to 2D max-projection images from WTC-11 and evaluates the learned representations on downstream tasks such as protein localization and cell-cycle prediction. Subcell~\cite{Subcell} learns representations from HPA~\cite{HPA} images using masked reconstruction and similarity-based objectives, and shows that image-derived features complement protein sequence and structure embeddings (e.g., ESM2~\cite{ESM2}), which are integrated via a second-stage multimodal model.
Existing image-based representation learning methods largely operate on 2D images or 2D projections of volumetric data. While 2D architectures can be extended to 3D via slice-wise processing \cite{uSegment3D}, we find that native 3D representation learning consistently outperforms both max-projection and slice-based approaches. Our model builds on SelfMedMAE~\cite{SelfMedMAE}, a 3D masked autoencoder originally developed for medical imaging, which we adapt to multi-channel fluorescence microscopy and further enhance with channel cross-attention and 3D frequency-domain regularization. Inspired by Subcell \cite{Subcell}, we further improve representation learning through alignment with protein sequence embeddings derived from ESM2~\cite{ESM2}. In contrast to Subcell, which learns a joint image–sequence representation in a second stage, we retain a purely image-based encoder and incorporate ESM2 during training via two mechanisms: (i) conditioning the decoder with sequence tokens for masked reconstruction, and (ii) aligning image and sequence embeddings with a symmetric InfoNCE~\cite{SimCLR} objective.











\begin{figure}[t]
    \centering
    \includegraphics[width=0.9\linewidth]{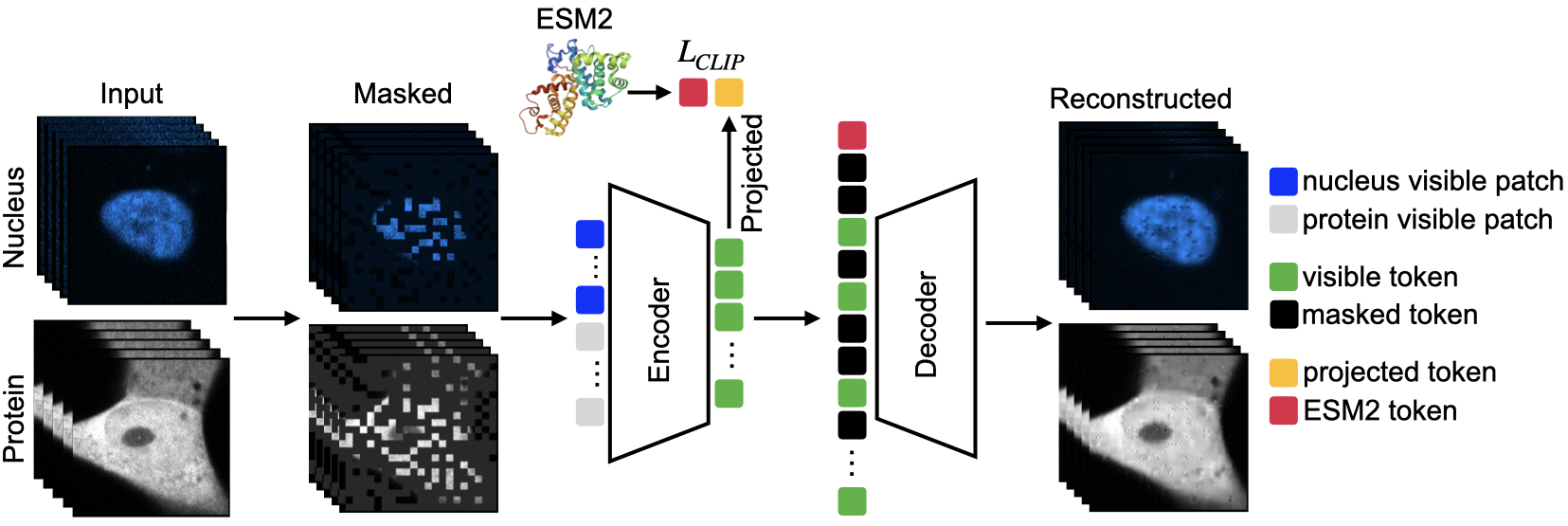}
    \caption{The MAE-3D$^{\star}$ model integrates protein representation by applying a contrastive $L_{\text{CLIP}}$ loss between projected image tokens from masked 3D OpenCell volumes and the corresponding ESM2 embedding~\cite{ESM2}. The ESM2 token is additionally fed to the decoder to guide reconstruction.}
    \label{fig:mae3d}
\end{figure}

\section{Methodology}

The proposed model is based on the Masked Autoencoder (MAE) framework in both 2D and 3D~\cite{MAE,SelfMedMAE}. The input is a $z$-stack volume $\mathbf{I} \in \mathbb{R}^{CZXY}$, where $C$, $Z$, $X$, and $Y$ denote the number of channels, depth, width, and height, respectively. For the 2D variant (MAE-2D$_{base}$), the volume is collapsed via a projection of maximum intensity along the z-axis, giving $\mathbf{I}_{2D}\! \in\!\mathbb{R}^{CXY}$. The $\mathbf{I}_{2D}$ image is patched into non-overlapping $p_x\!\times\!p_y$ patches. Patch embedding is performed using a convolutional layer
followed by 2D sinusoidal positional embeddings. A fraction $m$ of patches is randomly masked, and an encoder processes the visible patches. The latent representations are combined with mask tokens and passed to a decoder for reconstruction. For the 3D variant (MAE-3D$_{base}$), the depth dimension is retained, and the full $\mathbf{I}_{3D}$ volume is divided into non-overlapping patches of size $p_z \times p_x \times p_y$. Patch embedding is implemented via a 3D convolutional layer
followed by 3D sinusoidal positional embeddings. Masking and encoding follow the same procedure as in the 2D setting. The model is trained via mean squared error (MSE) loss over the masked patches. In both models, all channels are processed jointly.

\subsection{Channel Cross-Attention (CCA)}

The base models are extended to a dual-stream encoder–decoder to explicitly model inter-channel interactions~\cite{ChAdaViT}. Each channel is processed as a separate token stream with cross-attention between channels. A shared random masking pattern is applied identically across channels to prevent trivial reconstruction from unmasked positions in the complementary channel. Accordingly, patch embedding is performed independently for each channel. Within each channel, standard multi-head self-attention is applied over all visible tokens, capturing global spatial context. In addition, position-wise cross-attention~\cite{Attention} is introduced: a token at spatial index $i$ in channel $c_i$ queries only the token at the same spatial index in channel $c_j$\! ($i\!\neq\!j$). Since there is exactly one key per query position, the softmax operation degenerates to 1. Therefore, softmax is replaced with a sigmoid gating mechanism in attention computation~\cite{Attention}.
The decoder mirrors the encoder architecture. Models with channel cross-attention are denoted as CCA.

\subsection{FFT Loss}

An FFT-based~\cite{FocalFrequency} frequency loss is applied to the fully reconstructed output.
The loss is computed per channel $c$ on the reconstructed 2D (MAE-2D) or 3D (MAE-3D) image to preserve fine subcellular structures. While this frequency-domain loss was introduced for 2D reconstruction in~\cite{CAMAE}, we extend it to the 3D setting. The frequency-domain loss is defined as
\begin{equation}
\mathcal{L}_{\mathrm{FFT}}
= \frac{1}{2} \sum_{c}
L_1 \!\left(
\log\!\left(1 + \bigl\lvert \mathcal{F}(\hat{I}^{c}) \bigr\rvert\right),
\log\!\left(1 + \bigl\lvert \mathcal{F}(I^{c}) \bigr\rvert\right)
\right)
\end{equation}
where $I^c$ is the original and $\hat{I}^c$ is the reconstructed image at channel $c$, $\mathcal{F}$ denotes the $N$-dimensional discrete Fourier transform with orthonormal normalization, $|\cdot|$ is the magnitude spectrum, and $\log(1+\cdot)$ compresses the dynamic range. The $L_1$ distance is preferred over $L_2$ for robustness to frequency-domain outliers and only the magnitude spectrum is used. For MAE-2D, the transform is applied over the two spatial axes, whereas for MAE-3D it is computed over all three volumetric axes. The FFT loss is combined with the MSE loss using a weighting factor $w_{\mathrm{FFT}}$ which is set to zero during warm-up and linearly increased to its target value during a ramp-up phase, stabilizing early training and introducing the frequency constraint once reconstructions become structurally meaningful.

\subsection{Multimodal Alignment with ESM2}
\label{ESM2}

To model alignment between protein sequence information and cellular morphology, protein embeddings are injected into the MAE decoder via a contrastive image–sequence alignment objective~\cite{SimCLR}. The encoder architecture remains unchanged. The pretrained ESM2~\cite{ESM2} protein language model is kept frozen during MAE training and introduced only after stabilization of the reconstruction loss. The ESM2 embedding is projected to the decoder dimension through a learned linear layer, producing a single protein token. This token is inserted into each channel-specific decoder sequence alongside the encoder’s visible and mask tokens, using zero positional embedding (as it carries no spatial information) (see Figure~\ref{fig:mae3d}).
After decoding, the protein token is discarded prior to image reconstruction. This design allows masked patches to attend to protein context through the decoder’s self-attention mechanism. For cross-modal alignment, image embeddings from the encoder are projected to the ESM2 embedding dimension 
and a symmetric InfoNCE~\cite{SimCLR} objective with cosine similarity is applied:
\begin{equation}
\mathcal{L}_{\mathrm{CLIP}} 
= \frac{1}{2} \Big(
\mathrm{CE}(\tau \mathbf{E}_I \mathbf{E}_P^\top, \mathrm{diag}) 
+
\mathrm{CE}(\tau \mathbf{E}_P \mathbf{E}_I^\top, \mathrm{diag})
\Big)
\end{equation}
where $\mathbf{E}_I, \mathbf{E}_P$ 
denote the normalized image and protein projections, $\tau$ is a learnable temperature parameter,
where the target corresponds to matching diagonal pairs within the batch.
The cross-entropy (CE) terms enforce bidirectional alignment between image and protein modalities. The final loss becomes
\begin{equation}
    \mathcal{L} = \mathcal{L}_{\mathrm{MSE}} + w_{\mathrm{FFT}} \, \mathcal{L}_{\mathrm{FFT}} + w_{\mathrm{CLIP}} \mathcal{L}_{\mathrm{CLIP}}.
\end{equation}
Protein–image alignment is computed only on visible tokens, making the contrastive objective challenging and promoting robust representations.
Models incorporating CCA, FFT, and ESM2 are denoted as MAE-2D$^{\star}$ and MAE-3D$^{\star}$.

\noindent \textbf{Implementation Details}
MAE-2D and MAE-3D use a ViT~\cite{ViT} backbone with encoder/decoder dimensions 384/192, comprising 6 encoder and 4 decoder layers with 6 attention heads each. Patch sizes are $p_x=p_y=8$ in-plane and $p_z=10$ along the z-axis for MAE-3D. Models are trained for 10 epochs with 4-epoch linear warm-up; the FFT loss weight is set to $w_{\mathrm{FFT}}=0.1$ and ramped up over 2 epochs, with gradient clipping at 0.5. For protein alignment, a learning rate of $1\times10^{-5}$ and a projection dimension of 1280 (matching ESM2~\cite{ESM2}) are used. Training runs for 5 epochs with $w_{\mathrm{CLIP}}=1.0$ and temperature $\tau=0.07$, ramped up over 1 epoch; the decoder is frozen during the first warm-up epoch. All experiments are conducted on a single NVIDIA A100 80GB GPU.

\section{Experiments and Results}


The proposed method is evaluated on the OpenCell dataset~\cite{OpenCell}.
It is a large fluorescence microscopy dataset of human cells with endogenously tagged proteins. Each sample is a two-channel 3D $z$-stack (protein and nucleus). The dataset comprises 6,301 volumes from 1,310 proteins across 17 subcellular localization categories, with multiple cells per field of view. Single-cell crops are obtained using Cellpose~\cite{Cellpose} on 2D maximum-intensity projections and extracted from the corresponding 3D volumes, yielding 70,313 cells. All crops are standardized to $100 \times 176 \times 176$ pixels at $0.2\,\mu\text{m}$ isotropic resolution, where 100 denotes the number of z-slices. The learned representations are evaluated on two downstream tasks.
\textit{Protein localization:} Linear probing is performed over 17 compartments using frozen representations.  
\textit{Protein–protein interaction:} Binary classification of interacting protein pairs.
A two-layer MLP head (512$\rightarrow$128) is trained on frozen embeddings. All downstream models are trained for 100 epochs with a learning rate of $1 \times 10^{-4}$.
Protein sequences are retrieved from UniProt~\cite{UniProt2023} using Ensembl IDs provided on OpenCell. Sequence embeddings are obtained from the pretrained ESM2-650M~\cite{ESM2} model by mean pooling the final-layer token representations.
All experiments use five-fold cross-validation with protein-level splits to prevent data leakage.



\subsection{MAE-2D$_{\text{base}}$ and MAE-3D$_{\text{base}}$ Comparison}

\begin{figure}[t]
    \centering
    \includegraphics[width=0.85\linewidth]{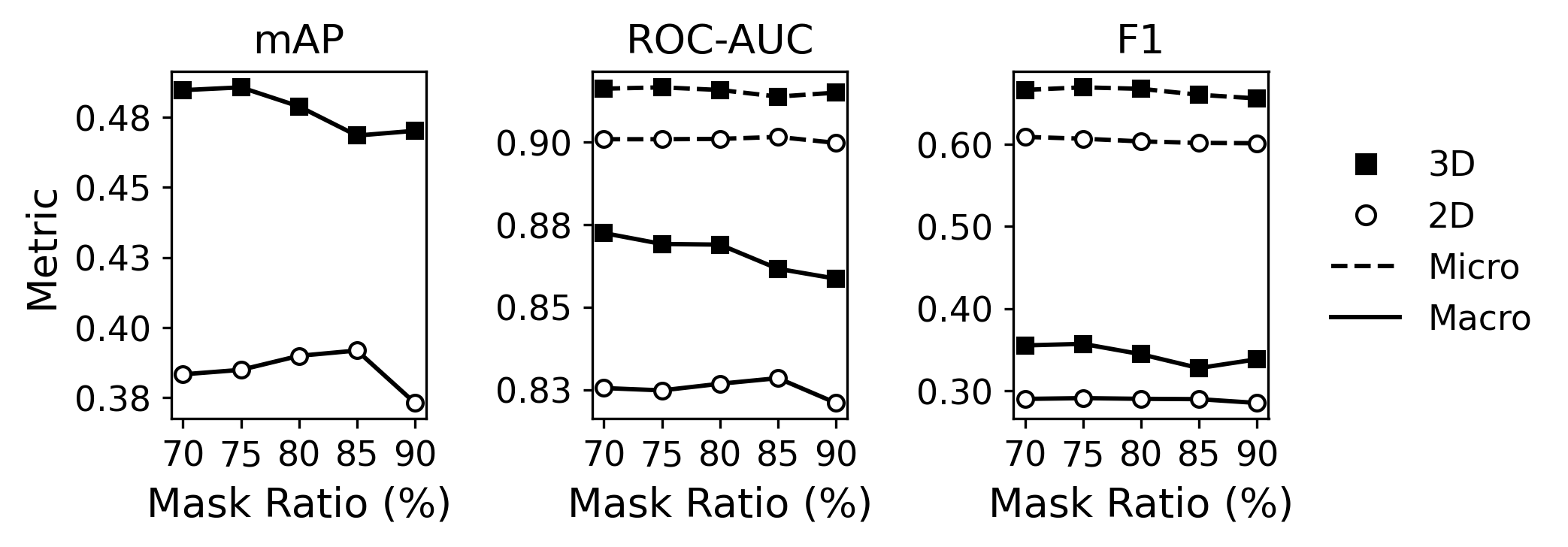}
    \caption{On the protein localization task, MAE-3D$_{\text{base}}$ outperforms MAE-2D$_{\text{base}}$ across all mask ratios and evaluation metrics on the OpenCell dataset.}
    
    \label{fig:mask_ratio}
\end{figure}

To compare base models and select an appropriate mask ratio for MAE pretraining, experiments were conducted on a single fold (Fold-1). Mask ratios of 70\%–90\% are evaluated for protein localization. As shown in Fig.~\ref{fig:mask_ratio}, MAE-3D$_{\text{base}}$ consistently outperforms MAE-2D$_{\text{base}}$ across all mask ratios, likely due to its access to full volumetric context. A mask ratio of 75\% achieves the best performance for MAE-3D$_{\text{base}}$ and is adopted for all subsequent experiments.
For a stricter comparison, both models are restricted to slices 45--55 per volume with 75\% masking. The 2D model is trained on individual slices per volume, requiring separate processing of each slice, which increases training time by approximately 10$\times$ for slices 45--55 and makes slice-based training substantially less computationally efficient than MAE-3D. Slice embeddings are averaged at inference to obtain a volume-level representation. Despite this, MAE-3D$_{base}$ still outperforms MAE-2D$_{base}$ (Table~\ref{tab:slice_comparison}). Furthermore, both models are evaluated on the protein--protein interaction task. MAE-3D$_{base}$ achieves substantially higher ROC-AUC compared to MAE-2D$_{base}$. Across proteins, the mean ROC-AUC improves from approximately 0.78 (2D) to 0.84 (3D). These results further confirm the benefit of modeling full volumetric context.



\begin{table}[t]
\centering
\caption{
MAE-3D$_{\text{base}}$ still outperforms MAE-2D$_{\text{base}}$ in protein localization when both models are restricted to slices 45--55.
}
\begin{tabular}{lccccc}
\toprule
Model 
& mAP 
& AUC$_{\text{macro}}$ 
& AUC$_{\text{micro}}$ 
& F1$_{\text{macro}}$ 
& F1$_{\text{micro}}$ \\
\midrule
MAE-2D$_{\text{base}}$ 
& 0.39 
& 0.81 
& 0.90 
& 0.30 
& 0.62 \\

MAE-3D$_{\text{base}}$ 
& \textbf{0.44} 
& \textbf{0.83} 
& \textbf{0.91} 
& \textbf{0.31} 
& \textbf{0.63} \\
\bottomrule
\end{tabular}
\label{tab:slice_comparison}
\end{table}

\begin{table}[t]
\centering
\scriptsize
\setlength{\tabcolsep}{4pt}
\renewcommand{\arraystretch}{1.15}
\caption{
On the protein localization task, MAE-3D$^{\star}$ outperforms MAE-2D$^{\star}$.
}
\label{tab:opencell_localization}
\begin{tabular}{lccccc}
\toprule
Model 
& mAP 
& AUC$_{\text{macro}}$ 
& AUC$_{\text{micro}}$ 
& F1$_{\text{macro}}$ 
& F1$_{\text{micro}}$ \\
\midrule
MAE-2D$^{\star}$ wo/ ESM2
& 0.54$\pm$0.03
& 0.89$\pm$0.01
& 0.94$\pm$0.01
& 0.48$\pm$0.02
& 0.70$\pm$0.01 \\

MAE-2D$^{\star}$
& 0.55$\pm$0.03
& 0.88$\pm$0.02
& 0.94$\pm$0.01
& 0.50$\pm$0.02
& 0.71$\pm$0.01 \\

MAE-3D$^{\star}$ wo/ ESM2
& 0.56$\pm$0.02
& 0.89$\pm$0.02
& 0.94$\pm$0.01
& 0.49$\pm$0.01
& 0.72$\pm$0.01 \\

MAE-3D$^{\star}$
& \textbf{0.62$\pm$0.02}
& \textbf{0.91$\pm$0.02}
& \textbf{0.95$\pm$0.01}
& \textbf{0.56$\pm$0.01}
& \textbf{0.74$\pm$0.01} \\
\bottomrule
\end{tabular}
\end{table}


\subsection{Ablation Study}

Table~\ref{tab:opencell_localization} assesses the impact of channel cross-attention (CCA), FFT loss, and ESM2 alignment on MAE-2D and MAE-3D for protein localization. Overall, MAE-3D$^\star$ achieves the best performance, with substantially larger gains from ESM2 than the 2D variant, indicating that sequence information better complements preserved 3D morphological context. On the protein--protein interaction (PPI) task, MAE-3D$^{\star}$ outperforms MAE-2D$^{\star}$, achieving a ROC--AUC of 0.86$\pm$0.03 compared to 0.84$\pm$0.04 for the 2D variant. The same trend holds without ESM2 (0.86$\pm$0.05 vs.\ 0.83$\pm$0.04), indicating that volumetric modeling provides consistent gains for interaction prediction.
\begin{figure}[t]
    \centering
    \includegraphics[width=0.95\linewidth]{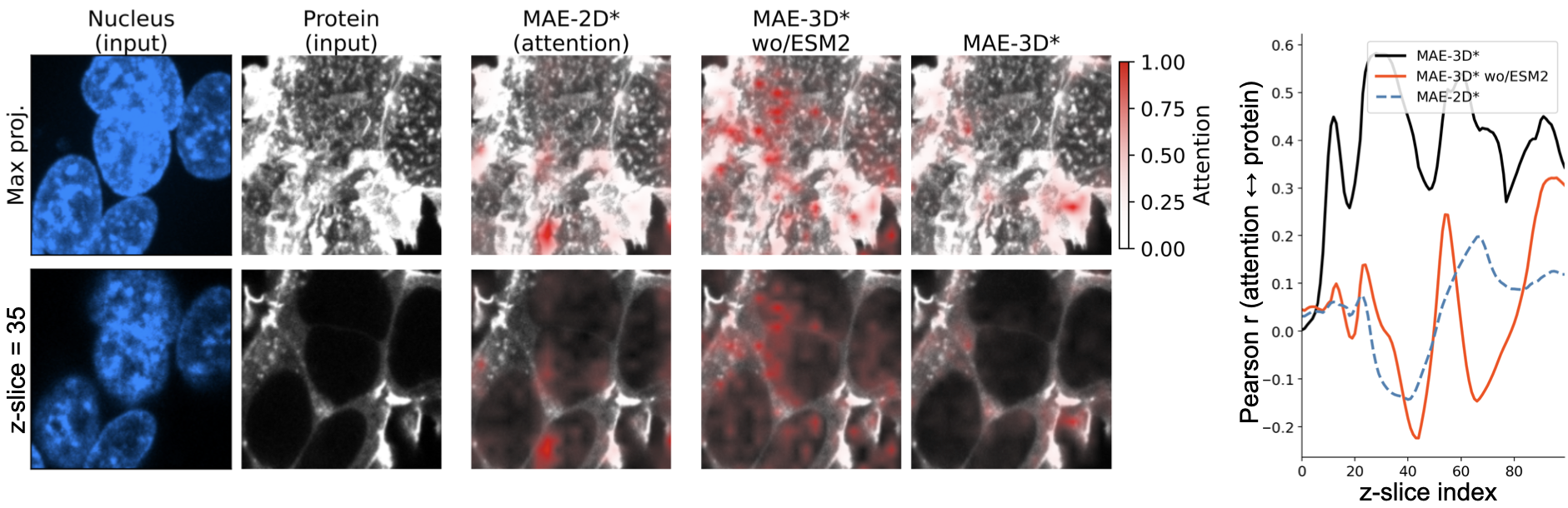}
    \caption{Integrating ESM2 leads to more precise attention over protein-specific regions, particularly across relevant $z$-stack slices in 3D. Attention is visualized for MAE-2D$^{\star}$, MAE-3D$^\star$ wo/ESM2, and MAE-3D$^\star$. Per-slice Pearson correlation ($r$) between attention and protein intensity is shown for each slice, where higher values indicate stronger spatial co-localization.
    }
    \label{fig:attn_vis}
\end{figure}
Figure~\ref{fig:attn_vis} shows attention maps for the exemplarily chosen protein ACTB, localized to the actin cytoskeleton and cytoplasm. With ESM2~\cite{ESM2}, attention concentrates on protein-relevant structures, whereas without it, focus spreads to non-specific regions, including the nucleus. The 3D model distributes attention consistently across relevant z-slices, reflecting volumetric awareness, while the 2D model concentrates on limited regions in the maximum-intensity projection, lacking depth specificity.

\subsection{Comparison with State-of-the-Art Methods}

On the OpenCell protein-protein interaction task, MAE-3D$^{\star}$ achieves the highest ROC-AUC (0.86 ± 0.03), outperforming Subcell \cite{Subcell} (0.85 ± 0.04) and DINO4Cell \cite{DINO4Cell} (0.84 ± 0.05). Performance without ESM2 (0.86 ± 0.05) is comparable with MAE-3D$^{\star}$, indicating that sequence-level alignment does not provide additional benefit for protein--protein interaction prediction.
For a fair comparison, Subcell and DINO4Cell embeddings are extracted for the same five-fold splits,
followed by linear probing, identical to our evaluation protocol. Our best-performing model (MAE-3D$^{\star}$) achieves the strongest results despite being trained on the comparatively small OpenCell dataset ($\sim$6k volumes), while Subcell and DINO4Cell are pretrained on large-scale HPA~\cite{HPA} microscopy datasets containing an order of magnitude more images. These findings underscore the effectiveness of volumetric modeling and cross-modal alignment, particularly in data-limited settings. As shown in Table~\ref{tab:opencell_sota}, MAE-3D$^{\star}$ achieves performance competitive with state-of-the-art methods on the protein localization task.


\begin{table}[t]
\centering
\scriptsize
\setlength{\tabcolsep}{3.5pt}
\renewcommand{\arraystretch}{1.15}
\caption{
Integrating ESM2 yields consistent gains for MAE-3D$^{\star}$, resulting in performance competitive with state-of-the-art methods and achieving the best micro-level scores.
}
\label{tab:opencell_sota}
\begin{tabular}{lccccc}
\toprule
Model 
& mAP 
& AUC$_{\text{macro}}$ 
& AUC$_{\text{micro}}$ 
& F1$_{\text{macro}}$ 
& F1$_{\text{micro}}$ \\
\midrule
Subcell
& 0.62$\pm$0.02
& \textbf{0.91$\pm$0.01}
& \textbf{0.95$\pm$0.01}
& 0.56$\pm$0.01
& 0.73$\pm$0.01 \\

DINO4Cell
& \textbf{0.63$\pm$0.01}
& 0.90$\pm$0.01
& \textbf{0.95$\pm$0.01}
& \textbf{0.57$\pm$0.02}
& 0.73$\pm$0.01 \\

MAE-3D$^{\star}$ wo/ESM2
& 0.56$\pm$0.02
& 0.89$\pm$0.02
& 0.94$\pm$0.01
& 0.49$\pm$0.01
& 0.72$\pm$0.01 \\

MAE-3D$^{\star}$
& 0.62$\pm$0.02
& 0.91$\pm$0.02
& \textbf{0.95$\pm$0.01}
& 0.56$\pm$0.01
& \textbf{0.74$\pm$0.01} \\
\bottomrule
\end{tabular}
\end{table}

\section{Conclusion}

We systematically compared 2D and 3D MAE-based models for representation learning in volumetric fluorescence microscopy. Across two tasks, native 3D models consistently outperformed 2D variants, demonstrating that preserving full volumetric context yields more discriminative representations. Integrating protein language embeddings (ESM2) further improved performance, with substantially larger gains for 3D models, highlighting a strong synergy between volumetric morphology and protein-level semantics. Overall, our findings emphasize the importance of native 3D modeling and multimodal alignment for robust foundation models in cellular imaging. Future work will explore learning a shared representation space across 2D and 3D datasets to better leverage protein sequence information, as well as extending the framework to subcellular segmentation with a dedicated decoder. \\

\begin{credits}


\subsubsection{\ackname} C.M. acknowledges support from the European Research Council (ERC; Grant Nos. 866411, 101113551, and 101213822), the High-tech Agenda Bayern, and the Deutsche Forschungsgemeinschaft (DFG, TRR359, Project No. 491676693). A.K. acknowledges computing time on the JUWELS Booster supercomputer operated by the Jülich Supercomputing Centre. L.G. acknowledges support from the Munich School for Data Science (MUDS).

\end{credits}

\bibliographystyle{splncs04}
\bibliography{mybibliography}

\end{document}